\documentclass[11pt]{article}
\usepackage{eacl2017}
\usepackage{times}
\usepackage{url}
\usepackage{latexsym}

\usepackage{amsmath}
\usepackage{graphicx}
\usepackage{multirow}
\usepackage{float}
\usepackage{caption}

\usepackage{subfig}

\usepackage{hhline}
\usepackage{pbox}
\usepackage[normalem]{ulem}
\eaclfinalcopy 
\def\eaclpaperid{120} 

\title{Improving a Strong Neural Parser with Conjunction-Specific Features}

\author{Jessica Ficler \\
 Computer Science Department \\
 Bar-Ilan University \\
 Israel \\
 {\tt jessica.ficler@gmail.com} \\\And
 Yoav Goldberg \\
 Computer Science Department \\
 Bar-Ilan University \\
 Israel \\
 {\tt yoav.goldberg@gmail.com} \\}

\date{}

\begin{document}
\maketitle
\begin{abstract}
While dependency parsers reach very high overall accuracy, some dependency relations are much harder than others. In particular,
dependency parsers perform poorly in coordination construction (i.e., correctly attaching the \textit{conj} relation).
We extend a state-of-the-art dependency parser with conjunction-specific features, focusing on the similarity between the conjuncts head words. Training the extended parser yields an improvement in \textit{conj} attachment as well as in overall dependency parsing accuracy on the Stanford dependency conversion of the Penn TreeBank.
\end{abstract}

\section{Introduction}
Advances in dependency parsing result in impressive overall parsing accuracy. For the most part, the advances are due to general improvements in parsing technology or feature representation, and do not explicitly target any specific language or syntactic construction. However, despite the high overall accuracy, parsers are still persistently wrong in attaching certain relations.
In the attachments predicted by BIST-parser \cite{kiperwasser2016simple}, the F1 score for the labels \textit{nn}, \textit{nsubj}, \textit{pobj}, and others is 95\% and above; while the F1 scores for \textit{advmod}, \textit{conj} and  \textit{prep} are 83.3\%, 82.5\% and 87.4\% respectively.
Conjunction holds the lowest F1 score, ignoring rare labels, \textit{dep} and \textit{punct}. Other parsers behave similarly.
Conjunction mistakes occurs also in simple sentences such as:

\vspace{5pt}
\noindent
\textit{(1) ``Those machines are still considered novelties, with keyboards only a munchkin could love and screens to match."} \\
\textit{(2) ``In the year-earlier period, CityFed had net income of \$ 485,000, but no per-share earnings."}
\vspace{5pt}

\noindent
BIST-parser \cite{kiperwasser2016simple} attaches \textit{screens} and \textit{love} instead \textit{screens} and \textit{keyboards} in (1); and \textit{earnings} and \textit{had} instead \textit{earnings} and \textit{income} in (2).

The parsers low performance on conjunction is disappointing given that conjunction is a common and important syntactic phenomena, appearing in almost 40\% of the sentences in the Penn TreeBank \cite{ptb}, 
as well constitutes 2.82\% of the Stanford dependency conversion of the Penn TreeBank \cite{de2008stanford} edges.

In this work we focus on improving \textit{conj} attachment accuracy by extending a dependency parser with features that specifically target the coordinating conjunction structures. Similar efforts were done for constituency parsing in previous work \cite{hogan2007coordinate,charniak2005coarse}.

As previously explored, conjuncts tend to be semantically related and have a similar syntactic structure \cite{shimbo2007discriminative,hara2009coordinate,hogan2007coordinate,ficler2016neural,charniak2005coarse,johnson1999estimators}.
For example: \textit{``for China and for India"}, \textit{``1.86 marks and 139.75 yen"}, \textit{``owns 33 \% of Moleculon’s stocks and holds 27.5 \% of Datapoint’s shares"}.
Such cases are common but still there are many cases where symmetry between conjuncts is less straightforward such as in (1), which includes the conjuncts \textit{``keyboards only a munchkin could love"} and \textit{``screens to match"}; and (2), which includes \textit{``net income of \$ 485,000"} and \textit{``no per-share earnings"}. 
For many cases of this type, the head words of the conjuncts are similar, e.g. \textit{(keyboards,screens)} in (1) and \textit{(income,earnings)} in (2).

We extend BIST-parser, the Bi-LSTM based parser by Kipwasser and Goldberg \shortcite{kiperwasser2016simple}, by adding explicit features that target the conjunction relation and focus on various aspects of symmetry between the potential conjuncts' head words. 
We show improvement in dependency  parsing scores and in \textit{conj} attachment.


\section{Symmetry between Conjuncts}
It is well known that conjuncts tend to be semantically related and often have a similar syntactic structure. This property of coordination was used as a guiding principle in
previous work on coordination disambiguation \cite{hara2009coordinate,hogan2007coordinate,shimbo2007discriminative,ficler2016neural}.
While these focus on symmetry between conjuncts in constituency structures, we use the symmetry assumption for the purpose of improving dependency parsing. 
Here is a simple example of dependency tree that include conjunction:
\begin{center}
\includegraphics[scale=0.5]{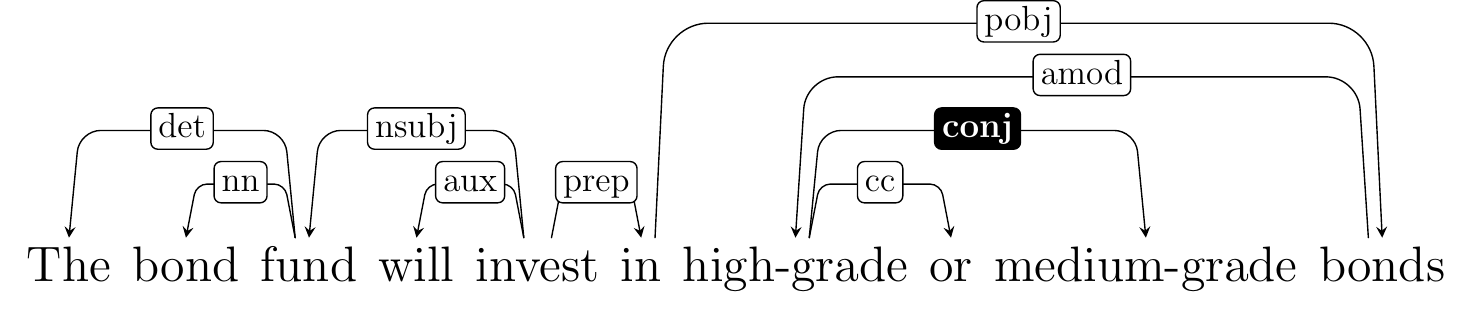}
\end{center}
The edge labeled with \textit{conj} connects the first conjunct head to the heads of the other conjuncts.
In more complex conjuncts, the subtrees under the nodes connected by \textit{conj} are often similar such as the following examples:
\begin{center}
\includegraphics[scale=0.55]{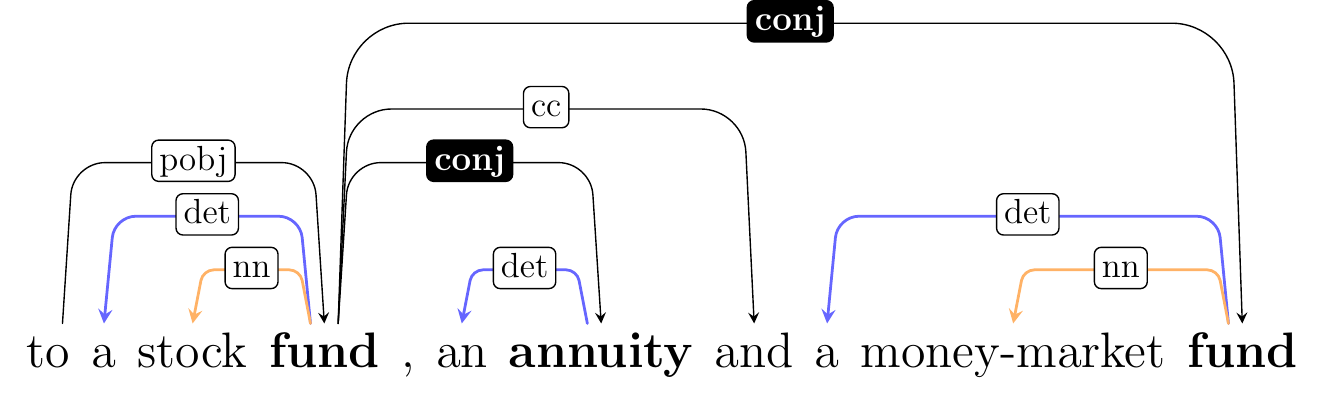}
\includegraphics[scale=0.46]{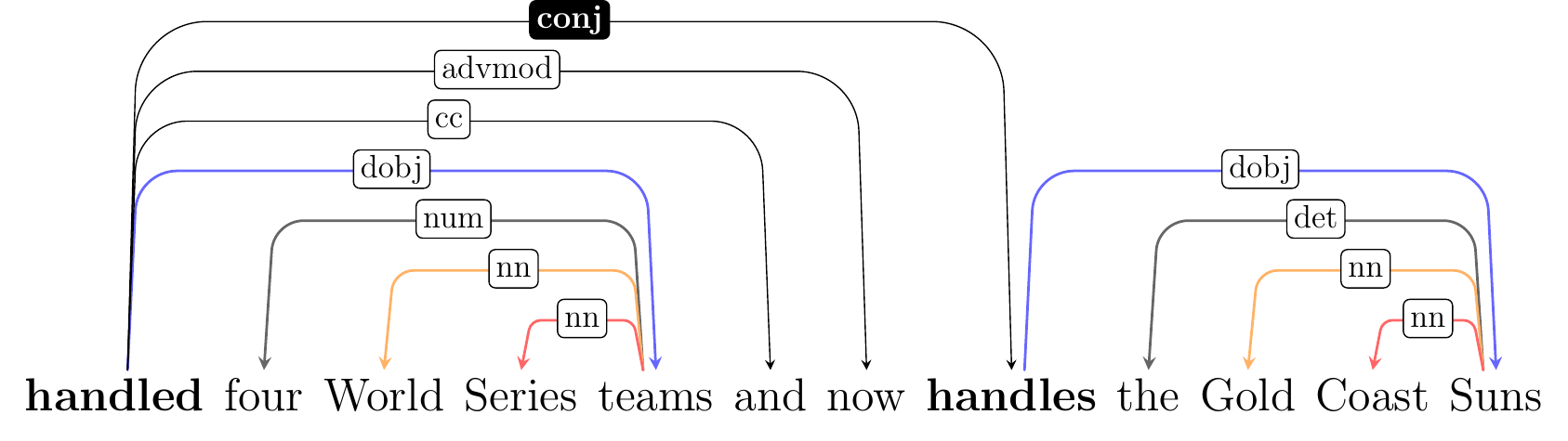}
\end{center}
However, there are also cases where the conjuncts structures are non-similar such as in:
\begin{center}
\includegraphics[scale=0.45]{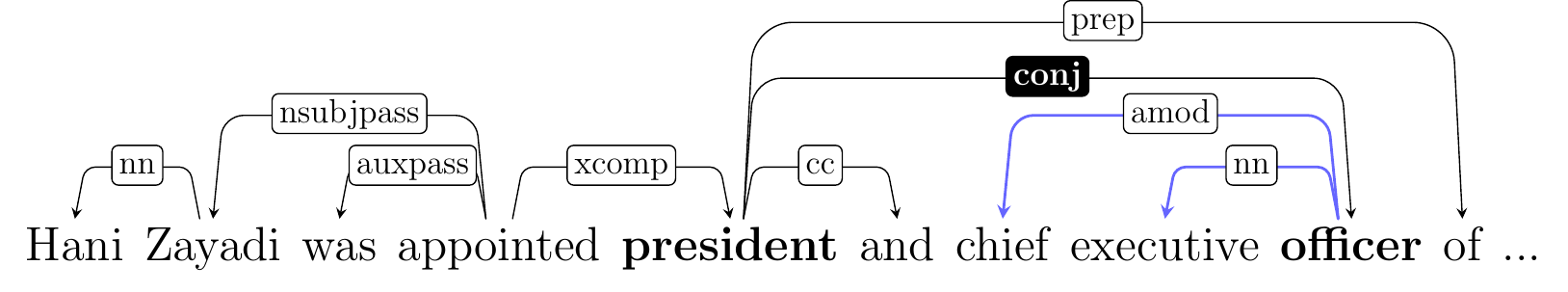}
\end{center}
Yet, some form of symmetry (or anti-symmetry) usually holds between the conjuncts head words.
Table \ref{tbl:conj_edjes_examples} lists the most common coordinated words in the PTB.


\begin{table}[t!]
\centering
\scalebox{0.8}{
\begin{tabular}{lclc} 
\multicolumn{4}{c}{(Head,Modifier)} \\
 \hline
1. & (\$,\$)  &  13. &(chairman,executive)     \\
2. &(\$,cents) & 14. &(on,on)     \\
3. &(president,officer)  &15. &(by,by)     \\
4. &(\%,\%)   &16. &(at,at)       \\
5. &(chairman,officer)        & 17. &(\$,\%)       \\
6. &(securities,exchange)     & 18. &(marks,yen)        \\
7. &(in,in)  &19. & (president,executive)  \\ 
8. &(standard,poor) &20. & (savings,association)    \\
9. &(to,to) &21. &    (chairman,president)      \\
10. &(buy,sell)   &22. &(inc.,inc.)       \\
11. &(for,for)&23. &(from,from)        \\
12. &(corp.,corp.)    &24. &(shares,\%)       \\
\end{tabular}}
\caption{The most common \textit{conj} attachments in the Penn TreeBank dependency conversion.  }
\label{tbl:conj_edjes_examples}
\end{table}

\section{Conjunction Features}
We suggest a set of features that are designed specifically for the conjunction relation, and target the symmetry aspect of the head words. The features look at a pair of head and modifier words, and are based on properties that appear frequently in conjunctions in the Stanford Dependencies version of the PTB. The features 
are summarized in Table \ref{tbl:local_feat}, and are detailed below:

\begin{table*}[t!]
\scalebox{0.72}{
\centering
\begin{tabular}{l|l|c|c} 
 & Description & Type & Examples   \\ 
 \hline
CAP  & Whether both words start with a capital letter  & boolean &  (Corp.,Inc.), (Poland,Hungary) \\
SUF  &  The length of the longer common suffix between the words & numeric & (men,women), (three-month,six-month )  \\
LEM  &  Whether the words lemmas are identical & boolean  & (say,said), (handled,handles) \\
SYM  & The cosine distance between the words embeddings  & numeric & (reported,said), (president,director)  \\
SENT-H&Whether the head word sentiment is positive, negative or neutral&1,-1, or 0&\multirow{2}{*}{(up,down), (confirmed,declined)}\\
   SENT-M&Whether the modifier word sentiment is positive, negative or neutral &1,-1, or 0 &\\

\end{tabular}}
\caption{Summary of the conjunction-specific features.}
\label{tbl:local_feat}
\end{table*}


\noindent
{\texttt{\bf CAP} --}
The case where both conjuncts head words start with a capital letter is much more common ($> 3\times$) than the case where only one of the head words starts with a capital letter. These cases are usually names of people, countries and organizations; and common phrases such as \textit{``Mac and Cheese"}. 
This property is rare in other labels except \textit{nn}.
We capture this property with a boolean feature that indicates whether both conjuncts head words start with a capital letter.

\noindent
{\texttt{\bf SUF} --}
In some of the conjunctions, the head words have a similar form, as in (codification, clarification), (demographic, geographic), (high-grade, medium-grade), (backwards, forwards). The cases where the longest common suffix between the words is at least 3 is 8\% in the case of \textit{conj} and much lower for the other labels. We capture this tendency using a numeric feature that indicated the length of the common suffix between the head words.

\noindent
{\texttt{\bf LEM} --}
Conjuncts heads often share the same lemma. These are usually different inflections of the same verb (e.g. sells,sold); or singular/plural forms of the same noun (e.g. table,tables).
This is also a tendency that is more common in \textit{conj} label than the other labels. 
We capture these, with a boolean feature indicating whether the lemmas of the conjuncts head words are identical. Lemmas are obtained using the NLTK \cite{bird2006nltk} interface to WordNet \cite{miller1995wordnet}.

\noindent
{\texttt{\bf SYM} --}
The conjuncts head words usually have a strong semantic relation. For example (fund, annuity), (same, similar), (buy, sell), (dishes, glass). \texttt{SYM} is a numeric feature that scores the similarity between the conjuncts heads words. The score is computed as the cosine-similarity between word embeddings of the head words (these embeddings are initialized with pre-trained vector from Dyer et al. (2015)).

\noindent
{\texttt{\bf SENT} --}
In some cases, both conjunct's head words sentiments are not neutral. Here are some examples from the PTB where both words are with positive sentiments: (enjoyable,easy), (complementary,interesting), (calm,rational); where both words are with negative sentiments: (slow,dump), (insulting,demeaning), (injury,death); and where one word is positive and the other is negative: (winners,losers), (crush,recover), (succeeded,failed). 
Having non-neutral sentiment for both words is not very common for \textit{conj} relation (2.3\% of the cases), but it much less common for the other relations. Therefore we add features that indicate the sentiment (positive, negative or neutral) for each of the coordinated words.
We use lists of positive and negative words from work on airline consumer sentiment \cite{breen2012mining}. 

\section{Incorporating conjunction features}
We incorporate the above features in the freely available BIST-parser ~\cite{kiperwasser2016simple}.
This parser is a greedy transition-based parser, using the archybrid
transition system \cite{kuhlmann2011dynamic}. 
At each step of the parsing process, the parser chooses one of $2*|labels|+1$ possible transitions: \textsc{Shift}, \textsc{Right}$_{(\text{rel})}$ and \textsc{Left}$_{\text{(rel)}}$. The \textsc{Left} and \textsc{Right} transitions add a dependency edge with the label \texttt{rel}. At each step, all transitions are scored, and the highest scoring transition is applied.

The Stanford Dependencies scheme specifies that the \textit{conj} relation appears as a right edge, and so it can only be produced by a \textsc{Right}$_{\text{(conj)}}$ transition. We compute a score $S_{\text{conj}}$ which is added to the score of the \textsc{Right}$_{\text{(conj)}}$ transition that was produced by the parser. 
$S_{conj}$ is computed by an MLP that receives a feature vector that is a concatenation of the original parser's features and the conjunction specific features. The scoring MLP and the parser are trained jointly.

\section{Experiments}

We evaluate the extended parsing model on the Stanford Dependencies \cite{de2008stanford} version of the Penn Treebank. We adapt BIST-parser code to run with the DyNet toolkit\footnote{\texttt{https://github.com/clab/dynet}} and add our changes.
We follow the setup of Kiperwasser and Goldberg \shortcite{kiperwasser2016simple}: (1) A word is represented as the concatenation of randomly initialized vector and pre-trained vector (taken from Dyer et al. (2015)); (2) The word and POS embeddings are tuned during training; (3) Punctuation symbols
are not considered in the evaluation;
(4) The hyper-parameters values are as in Kiperwasser and Goldberg paper ~\shortcite{kiperwasser2016simple}, Table 2;
(5) We use the same seed and do not perform hyper-parameter tuning. 
We train the parser with the conjunction features 
for up to 10 iterations, and
choose the best model according to the LAS accuracy
on the development set.

\paragraph{General Parsing Results}
Table \ref{tbl:results} compares our results to the unmodified BIST parser. The extended parser achieves 0.1 points improvement in UAS and 0.2 points in LAS comparing to Kiperwasser and Goldberg \shortcite{kiperwasser2016simple}. This is a strong baseline, which so far held the highest results among greedy transition based parsers that were trained on the PTB only, including e.g. the parsers of Weiss et al \shortcite{weiss2015structured}, Dyer et al \shortcite{dyer2015transition} and Ballesteros et al \shortcite{ballesteros2016training}. Stronger absolute parsing numbers are reported by Andor et al \shortcite{andor2016globally} (using a beam); and Kuncoro et al \shortcite{kuncoro2016distilling} and Dozat and Manning \shortcite{dozat2016deep} (using an arc-factored global parsers). All those parsers rely on broadly the same kind of features, and while we did not test this, it is likely the conjunction features would benefit them as well.\footnote{A reviewer of this work suggested that our baseline model is oblivious to the word's morphology, and that a neural parsing architecture that explicitly models the words' morphology through character-based LSTMs, such as the model of \cite{ballesteros2015improved}, could capture some of the information in our features automatically, and thus would be a better baseline. While we were skeptical, we tried this suggestion, and found that it indeed does not change the results in a meaningful way.}

\paragraph{Parsing Results for \textit{conj} Label}
We evaluate our model specifically for \textit{conj} label, and compare to the results achieved by the parser without the conjunction features.
We measure Rel (correctly identifying modifiers that participate in a conj relation, regardless of correctly attaching the parent) and Rel+Att (correctly identifying both the head and the modifier in a conj relation).
The results are in Table \ref{tbl:conj_results}. The improvement in Rel score is relatively small while there is an improvement of 1.1 points in Rel+Att F1 score, suggesting that the parser was already effective at identifying the modifiers in a conj relation and that our model's benefit is mainly on attaching the correct parent node.

\paragraph{Analysis}
We would like to examine to what extent the improvement we achieve over 
Kiperwasser and Goldberg \shortcite{kiperwasser2016simple} on \textit{conj} attachments corresponds to the coordination features we designed.
To do that,
we analyze the \textit{conj} cases in the dev-set that were correctly predicted by our model and were not predicted by the original BIST-parser and vice versa.
The following table shows the percentage of cases where conjunction features appear in each of these lists:

\begin{center}
\scalebox{0.82}{
\begin{tabular}{l|c|c} 
Features &  $+$Our,$-$\small{K\&G} & $-$Our,$+$\small{K\&G}  \\ 
 \hline
\texttt{LEM}+\texttt{CAP}+\texttt{SUF} & 7.5 &0\\
\texttt{LEM}+\texttt{SUF} &3&0 \\
\texttt{SENTIMENT}+\texttt{SUF} &1.5 &0 \\
\texttt{LEM}/\texttt{CAP}/\texttt{SENTIMENT}/\texttt{SUF}&29.9&24  \\
Total &41.9 &24 \\
\end{tabular}}
\end{center}
The percentage of cases that include conjunction features is much higher in the list of cases that were correctly predicted only by our model. More than that, there are no cases that include more than one conjunction feature in the list of cases that were correctly predicted only by BIST-parser \cite{kiperwasser2016simple}.

The above table does not include the \texttt{SYM} feature since unlike the other features there is no absolute way to determine whether the feature takes place on a specific example. To give a sense of the contribution of the \texttt{SYM} feature, we show some examples where our model attaches a \textit{conj} label between similar words, while the unmodified BIST parser attaches \textit{conj} parent which is clearly less similar to the modifier (The word in bold is the attached modifier; the word marked with continuous line is the node's parent in our prediction; the word marked with dashed line is the node's parent in BIST's prediction):

\begin{itemize}
\setlength{\itemsep}{0pt}
\setlength{\parskip}{0pt}
\item Koop, who \dashuline{rattled} \underline{liberals} and \textbf{conservatives} alike with his outspoken views on ...
\item ... dropped in response to \underline{gains} in the stock \dashuline{market} and \textbf{losses} in Treasury securities.
\item Died: Cornel \dashuline{Wilde}, 74 ,\underline{actor} and \textbf{director} ,in Los Angeles ,of leukemia ... 
\item ... investment firms \dashuline{advising} clients to \underline{boost} their stock holdings and \textbf{reduce} the ,,.
\end{itemize}

In the cases that were correctly predicted by BIST-parser only, we could not find examples where the words in the correct attachment are clearly more similar than the attachment predicted by our model. We could find a few examples where both models attached words that are similar, such as: 
\begin{itemize}
\setlength{\itemsep}{0pt}
\setlength{\parskip}{0pt}
\item ML \& Co.'s net income \underline{dropped} 37\%, while BS Cos. posted a 7.5\% gain in net, and PG Inc.'s profit \dashuline{fell}, but would have \textbf{risen} ...
\item The closely watched \underline{rate} on federal  \dashuline{funds}, or overnight \textbf{loans} between banks, slid to...
\end{itemize}

\begin{table}[t!]
\scalebox{0.82}{
\centering
\begin{tabular}{l|c|c} 
 System & UAS & LAS\\ 
 \hline
Kiperwasser16  & 93.9 & 91.9 \\
Kiperwasser16 + conjunction features &  \bf 94 & \bf 92.1  \\
\end{tabular}}
\caption{Parsing scores on the PTB test-set (Stanford Dependencies).}

\label{tbl:results}
\end{table}

\begin{table}
\scalebox{0.82}{
\centering
\begin{tabular}{l|c|c} 
&Kiperwasser16 &  \pbox{20cm}{Kiperwasser16 + \\ conjunction features}\\
\hline
 Rel R &92.5 & \bf 92.9\\
 Rel P &\bf 91.6&91.5\\
 Rel F1& 92 & \bf 92.2\\
 \hline \hline
 Rel+Att R & 83 & \bf 84.2\\
  Rel+Att P & 82.1 &\bf 83 \\
  Rel+Att F1& 82.5 & \bf 83.6\\ 
\end{tabular}}
\caption{Test-set results for \textit{conj} label only.}

\label{tbl:conj_results}
\end{table}

\section{Conclusions}
While most recent work in parsing attempt to improve results using "general" architectures and feature sets, targeted feature engineering is still beneficial. We demonstrate that a linguistically motivated and data-driven feature-set for a specific syntactic relation (coordinating conjunction) improves a strong baseline parser.

The features we propose explicitly model the symmetry between the head words in coordination constructions. While we demonstrated their effectiveness in a greedy transition-based parser, the information our features capture is not currently captured also by other dependency parsing architectures (including first-order graph based parsers, higher-order graph-based parsers, beam-based transition parsers). These features will be straightforward to integrate into such parsers, and we expect them to be effective for them as well.

\section*{Acknowledgments}
This work was supported by The Israeli Science Foundation (grant number 1555/15)
as well as the German Research Foundation via the
German-Israeli Project Cooperation (DIP, grant DA 1600/1-1).

\bibliography{eacl2017}
\bibliographystyle{eacl2017}

\end{document}